\documentclass[ngerman,english]{bvm}

\bvmdef\articlenumber{3067}

\bvmdef\type{S}
\title{Coronary Plaque Analysis for CT Angiography Clinical Research}
\titlerunning{Coronary Plaque Analysis for CTA Research}

\author{Felix~Denzinger$^{1,2,5}$, Michael~Wels$^{2,5}$, Christian~Hopfgartner$^3$, Jing~Lu$^2$, Max~Sch\"obinger$^2$, Andreas~Maier$^{1,4}$, Michael~S\"uhling$^2$} 

\authorrunning{Denzinger~Wels~et~al.}

\institute{%
	$^1$Pattern~Recognition~Lab,~Friedrich-Alexander-Universit\"at~Erlangen-N\"urnberg,~Erlangen\\
	$^2$Computed~Tomography,~Siemens~Healthineers,~Forchheim,~Germany\\
    $^3$ISO-Gruppe,~Nuremberg,~Germany\\
    $^4$Machine~Intelligence,~Friedrich-Alexander-Universit\"at~Erlangen-N\"urnberg,~Erlangen\\
    $^5$Contributed~equally~to~this~work. 
}
    
\email{felix.denzinger@fau.de}

\begin{document}
	
	%
	\selectlanguage{english}
	
	\maketitle
	
	\begin{abstract}
		The analysis of plaque deposits in the coronary vasculature is an important topic in current clinical research.
		From a technical side mostly new algorithms for different sub tasks -- e.g. centerline extraction or vessel/plaque segmentation -- are proposed. 
		However, to enable clinical research with the help of these algorithms, a software solution, which enables manual correction, comprehensive visual feedback and tissue analysis capabilities, is needed. 
		Therefore, we want to present such an integrated software solution.\footnote{A MeVisLab-based implementation of our solution is available as part of the Siemens Healthineers syngo.via Frontier and OpenApps research extension.}
		It is able to perform robust automatic centerline extraction and inner and outer vessel wall segmentation, while providing easy to use manual correction tools.
		Also, it allows for annotation of lesions along the centerlines, which can be further analyzed regarding their tissue composition.
		Furthermore, it enables research in upcoming technologies and research directions: it does support dual energy CT scans with dedicated plaque analysis and the quantification of the fatty tissue surrounding the vasculature, also in automated set-ups.
	\end{abstract}
	
	\section{Introduction}
	%
	
	Cardiovascular diseases (CVDs) are the leading cause of natural death \cite{3067-01}. Therefore, the interest in clinical research in this area is high. Most CVDs are related to atherosclerotic plaque deposits and the composition of these plaques plays an important role in the patient outcome and risk stratification \cite{3067-02}. 
	Modern cardiac computed-tomography angiography (CCTA) provides means to assess the morphology of plaque deposits in the coronary arteries.
	
	However, exact quantitative analysis of these deposits is tedious without appropriate tools. If these tools are not available, clinical researchers are not able to efficiently evaluate plaque deposits on larger patient cohorts, hindering them from contributing to progress in evidence-based medicine. Also, clinical studies need to be reproducible, which is easier to achieve -- especially for studies with the focus on plaque analysis, with a high amount of processing steps -- when most of the pipeline is automated. This also leads to a reduction of the amount of inter- and intra-observer variance.
	
    To overcome these hurdles, we want to present a software solution and an associated semi-automatic general workflow for quantitative and semantic coronary artery plaque analysis from CCTA data with a focus on the underlying algorithms. As a semi-automatic approach it allows more efficient, accurate and reproducible plaque analysis. By allowing appropriate user-interaction at each processing step, the system is prevented from generating flawed final results.
    
    Furthermore, two major topics in upcoming and current research in the area of CCTA are supported: dual energy (DE) CT scans are automatically detected as such, registered onto each other and the composition of the plaque deposits can be analyzed using information from two energy spectra. Also, the composition of the fatty tissue surrounding the vasculature was shown to be correlated with inflammation and consequently plaque aggregation and cardiac death \cite{3067-03}. In our software solution we additionally enable automated segmentation of the regions of interest for this analysis. 
    
    Since we present a tool-chain for clinical research this paper explains the individual elements of this software solution and highlights some related already published applications in clinical research.
    %
    \begin{figure}[t]
        \centering
        \caption{Processing steps of our solution: (1) heart isolation, (2) centerline extraction, (3) ROI definition and segmentation, (4) plaque deposition analysis and annotation.}
    	\label{3067-fig-01}
        \includegraphics[width=0.9\textwidth]{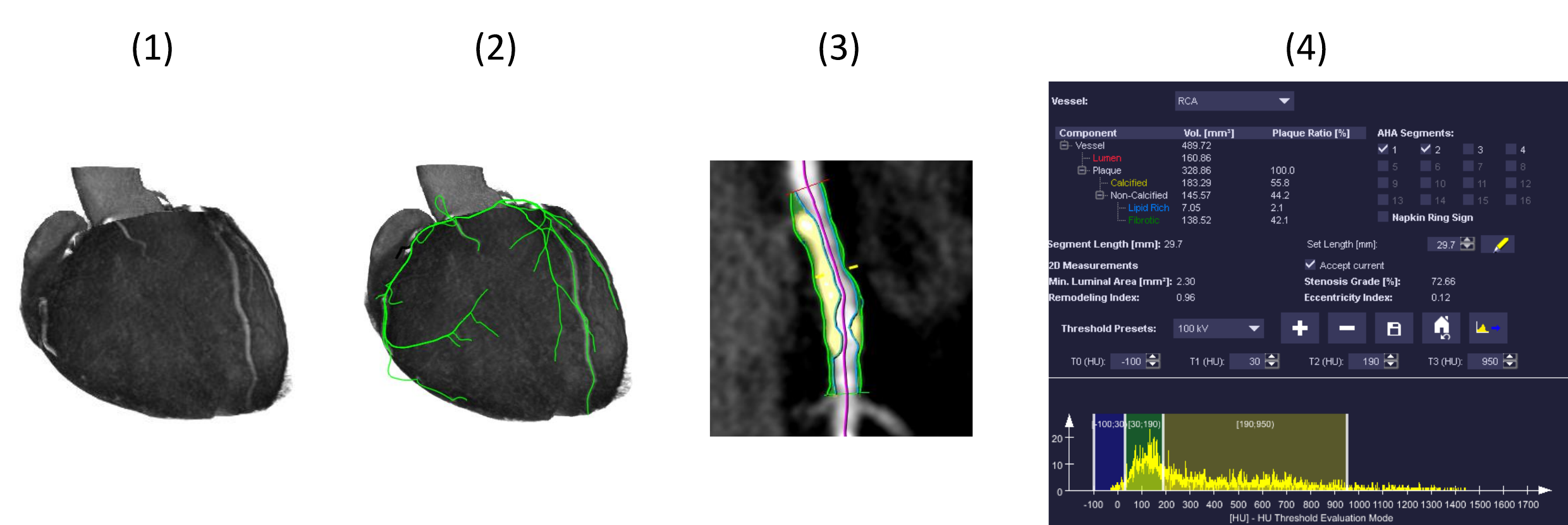}
    \end{figure} 
	\section{Material and Methods}
	\subsection{General workflow}
	
	Starting from a CCTA volume as input, our tool-chain consists of the following steps: fully automatic heart isolation and coronary artery centerline detection, including centerline correction, then -- if needed -- tools for a manual correction of these pre-processing results, next, selection of the vessel branch and section of interest with a consecutive fully automatic vessel wall segmentation, and finally an analysis of the plaque composition and quantification using manually set Hounsfield unit (HU) thresholds. The workflow of our prototype software solution is depicted in Fig.~\ref{3067-fig-01}. As soon as the volume is loaded the heart and its anatomies are fully-automatically detected with the method described by Zheng et al.~\cite{3067-04}, which utilizes marginal space learning and steerable features and is described to be robust and fast ($\approx4$ seconds per volume). The heart isolation is needed for further fully-automatic coronary artery centerline extraction. The method used for this step is described by Zheng et al. \cite{3067-05} as a both model- and data-driven approach, which takes the prior information about vessel specific regions of interest (ROIs) in order to allow robust centerline extraction also for occluded vessels. This method is as of today still the best performing fully-automatic approach on the Rotterdam Coronary Centerline Extraction Challenge leaderboard \cite{3067-14}. If the result of the centerline extraction step is not satisfactory it can be corrected by a simple interactive tool and additional centerlines can be manually drawn or created considering image evidence with a single seed point. 
	
	The vessel section to be analyzed can be selected by clicking on the start and end point on the centerline. The created markers can be shifted along the centerline if one is not satisfied with the initial selection. 
	Next, the segmentation of both the inner and outer vessel wall within the area of interest can be triggered.
	For the inner wall segmentation the approach described in \cite{3067-06} is utilized. It is based on ray-casting and the analysis of a subsequent Markov random field graph with convex priors. The described approach is known to be robust and accurate on a subvoxel level as its the current leader on the Rotterdam Coronary Lumen Segmentation Challenge leaderboard \cite{3067-15}.
	
	For the outer vessel wall an adaptive self-learning edge-model using a combination of 3D and 2D active contour models is used \cite{3067-07}. By adaptation of the threshold for this approach the outer vessel wall segmentation can be widened or narrowed in real-time on cross-sectional views and then applied for the whole region of interest. Especially the outer wall segmentation is reader-dependent, and often not clearly defined, since there is a lack of contrast enhancement due to soft plaques within the arterial walls. Therefore, this is a good compromise between robustness and flexibility allowing a fast adaptation.
	
	In order to correct or adapt the resulting inner and outer wall segmentation a quasi-real-time interactive method is used \cite{3067-08}. By describing the inner and outer vessel wall as implicit surfaces with radial basis functions, manual corrections in the curved or axial 2D views of the vessel can be directly propagated to the 3D surface and corrected accordingly.
	
	\subsection{Plaque Analysis}
	The resulting plaque region then can be analyzed, e.g. with respect to its composition. Since the exact thresholds for different tissue types vary for different tube voltages, the thresholds between the mostly used tissue classes -- lipid-rich, fibrotic and calcified -- can be adapted directly in the histogram of the HU values.
	Additionally, this HU histogram of the ROI can be directly exported to enable further analysis of the value distribution.
	Moreover, an important factor in assessing coronary plaques are high risk plaque features. These include the degree of stenosis, positive remodeling, low HU attenuation and the so-called napkin ring sign. The degree of stenosis and remodeling can be determined for each individual section of the annotated lesion by taking the proximal and distal markers as weighted reference for each position. While low HU attenuation plaques can be derived from the plaque composition histogram the napkin ring sign can be annotated manually. 
	
	In the last decade DE scanners were introduced, which can provide a better tissue contrast in some areas due to two different energy spectra. According to literature \cite{3067-09}, calcified plaques get well detected by single energy CT scans but lipid-rich and fibrotic tissue are harder to differentiate. Since this differentiation is especially important for the classification of high-risk plaques, the interest in plaque analysis with DE scans is high.
	Therefore, our presented software solution does support DE scans. 
	Solely a prior registration of the two DE scans is needed to enable the same processing pipeline as described above. The plaque deposits can then be first examined regarding whether they are calcified or not with a single energy volume and then their composition can be further analyzed combining information from the two scans and thresholding the DE index. This follows the ideas proposed by \cite{3067-10}.
	
	Another rising topic in CCTA clinical research is the analysis of perivascular fat \cite{3067-03}. The HU distribution of the fat surrounding the coronary vasculature was shown to be correlated with inflammation leading to plaque aggregation. To enable research in this direction, we support the creation of ROIs which expand radially to the vessel dependent on either the inner or outer vessel wall. The radius of these ROIs can be freely set. Since in literature standardized ROIs exist, we provide an automated mode which creates one ROI for each main branch according to the definition in the literature.

	\section{Results}
	\begin{table}
	    \centering
	    
        \caption{Comparison to other available approaches. The features of alternate solutions are collected from public sources and are to the best of our knowledge. All available approaches contain tools for semi-automated vessel segmentation, centerline correction, plaque composition analysis and segmentation correction.}
        \label{3067-tab-01}
        \begin{tabular*}{\textwidth}{ c@{\extracolsep\fill} c c c c c c }
        \hline
          & \rotatebox[origin=c]{90}{Ours} & 
          \rotatebox[origin=c]{90}{\begin{tabular}[c]{@{}c@{}}Cedars Sinai \\AutoPlaq\footnotemark{}\end{tabular} }
        & \rotatebox[origin=c]{90}{\begin{tabular}[c]{@{}c@{}}Medis \\QAngio CT\footnotemark{}\end{tabular}}  
        & \rotatebox[origin=c]{90}{\begin{tabular}[c]{@{}c@{}}Canon CT\\ SUREPlaque\footnotemark{}\end{tabular} }
        & \rotatebox[origin=c]{90}{\begin{tabular}[c]{@{}c@{}}GE VesselIQ\\ Xpress \footnotemark{}\end{tabular}}
        & \rotatebox[origin=c]{90}{\begin{tabular}[c]{@{}c@{}}Philips \\CT CCA\footnotemark{}\end{tabular}}
        
        \\ \hline
        \begin{tabular}[c]{@{}c@{}}Automated \\ Pre-processing\end{tabular}          & Yes  & No & Yes   & Semi     & Yes & Yes  \\ 
        \begin{tabular}[c]{@{}c@{}}Dual Energy\\ Support\end{tabular}                & Yes  & No      & No        & No    & No   & No     \\ 
        \begin{tabular}[c]{@{}c@{}}Perivascular Tissue\\ Analysis\end{tabular}       & Yes  & Yes     & No        & No     & No  & No     \\ 
        \begin{tabular}[c]{@{}c@{}}Automated Segment \\ Labeling\end{tabular}       & No  & Unknown     & Yes        & No      & Yes   & Yes   \\ \hline

        \end{tabular*}
        
    \end{table}
    Since multiple sources provide such a software solution, we want to briefly differentiate our solution to other available tools. An overview of this can be seen in Table~\ref{3067-tab-01}. While many other approaches do support standard pre-processing, segmentation and correction capabilities, our solution provides DE support, which no other solution to the best of our knowledge does. Also only the solution of Cedars Sinai does include perivascular tissue analysis but not in the automated setting we do.

	Due to the nature of this paper, we will not provide quantitative results. In total 16 publications utilized our software solution up until today. Due to space restrictions, we are only able to highlight some of these here.
	First, we want to present work of Tesche et al. \cite{3067-11}. They used our solution to correlate different CCTA derived plaque characteristics with the questions whether lesions are intra-operatively determined to be haemodynamically significant. In their study they examined 37 lesions of 37 patients but they have increased the number of patients in later studies. 
	Furthermore, the research of Ratiu et al. \cite{3067-12} aims towards analyzing the lesion geometry as an additional indicator for high risk plaque segments. 
	Morariu et al. \cite{3067-13} are currently conducting a trial examining also the characteristics of plaque deposits after initial myocardiac infarction. They plan to collect up to 100 patients with major adverse cardiac events as potential endpoint.

    \footnotetext[2]{https://www.cedars-sinai.org/research/labs/dey.html}
	\footnotetext[3]{https://medisimaging.com/apps/plaque-burden/}
	\footnotetext[4]{https://www.vitalimages.com/product-information/ct-sureplaque/}
	\footnotetext[5]{https://www.gehealthcare.co.uk/products/advanced-visualization/all-applications/autobone-vesseliq-xpress}
	\footnotetext[6]{https://www.philips.co.uk/healthcare/product/HCAPP006/-ct-comprehensive-cardiac-analysis-cca-}

	\section{Discussion}
	In this paper we described a software solution which enables plaque and tissue analysis-related clinical research in the field of coronary artery diseases.
	In the workflow of this prototype all necessary processing steps for coronary plaque analysis exist and easy manual correction is able for all steps. The majority of fields of research which are currently in the focus are covered by our solution. These include the support of DE scans and the analysis of perivascular tissue, which are hardly supported by most other available solutions. The fully automatic labelling of the branch segments will be part of further improvement to the software.
	We presented already conducted clinical studies with our software solution, which proofs the applicability of described workflows and components in this important clinical research field.
	
	\bibliographystyle{bvm}
	
	\bibliography{3067}

\begin{thebibliography}{10}

\bibitem{3067-01}
Mendis S, Davis S, Norrving B.
\newblock Organizational update: the {World Health Organization} global status
  report on noncommunicable diseases 2014.
\newblock Stroke. 2015;46(5):e121--e122.

\bibitem{3067-02}
Naghavi M.
\newblock From vulnerable plaque to vulnerable patient: a call for new
  definitions and risk assessment strategies. {Part II}.
\newblock Circulation. 2003;108:1772--1778.

\bibitem{3067-03}
Antonopoulos AS, et~al.
\newblock Detecting human coronary inflammation by imaging perivascular fat.
\newblock Sci Transl Med. 2017;9(398).

\bibitem{3067-04}
Zheng Y, Barbu A, Georgescu B, et~al.
\newblock {Four-chamber heart modeling and automatic segmentation for 3-D
  cardiac CT volumes using marginal space learning and steerable features}.
\newblock {IEEE} Trans Med Imaging. 2008;27(11):1668--1681.

\bibitem{3067-05}
Zheng Y, Tek H, Funka-Lea G.
\newblock Robust and accurate coronary artery centerline extraction in {CTA} by
  combining model-driven and data-driven approaches.
\newblock In: MICCAI. Springer; 2013.  p. 74--81.

\bibitem{3067-14}
Schaap M, Metz CT, van Walsum T, et~al.
\newblock Standardized evaluation methodology and reference database for
  evaluating coronary artery centerline extraction algorithms.
\newblock Med Image Anal. 2009;13(5):701--714.

\bibitem{3067-06}
Lugauer F, Zheng Y, Hornegger J, et~al.
\newblock Precise lumen segmentation in coronary computed tomography
  angiography.
\newblock In: International MICCAI Workshop on Medical Computer Vision.
  Springer; 2014.  p. 137--147.

\bibitem{3067-15}
Kiri{\c{s}}li H, et~al.
\newblock Standardized evaluation framework for evaluating coronary artery
  stenosis detection, stenosis quantification and lumen segmentation algorithms
  in computed tomography angiography.
\newblock Med Image Anal. 2013;17(8):859--876.

\bibitem{3067-07}
Grosskopf S, Biermann C, Deng K, et~al.
\newblock Accurate, fast, and robust vessel contour segmentation of {CTA} using
  an adaptive self-learning edge model.
\newblock In: Medical Imaging 2009: Image Processing. vol. 7259. International
  Society for Optics and Photonics; 2009.  p. 72594D.

\bibitem{3067-08}
Wels M, Lades F, Hopfgartner C, et~al.
\newblock Intuitive and Accurate Patient-Specific Coronary Tree Modeling from
  Cardiac Computed-Tomography Angiography.
\newblock In: The 3rd interactive {MIC Workshop}; 2016.  p. 86--93.

\bibitem{3067-09}
Danad I, {\'O}~Hartaigh B, Min JK.
\newblock Dual-energy computed tomography for detection of coronary artery
  disease.
\newblock Expert Rev Cardiovasc Ther. 2015;13(12):1345--1356.

\bibitem{3067-10}
Barreto M, Schoenhagen P, Nair A, et~al.
\newblock Potential of dual-energy computed tomography to characterize
  atherosclerotic plaque: ex vivo assessment of human coronary arteries in
  comparison to histology.
\newblock J Cardiovasc Comput Tomogr. 2008;2(4):234--242.

\bibitem{3067-11}
Tesche C, et~al.
\newblock {Coronary CT angiography derived morphological and functional
  quantitative plaque markers correlated with invasive fractional flow reserve
  for detecting hemodynamically significant stenosis}.
\newblock J Cardiovasc Comput Tomogr. 2016;10(3):199--206.

\bibitem{3067-12}
Ratiu M, et~al.
\newblock Impact of coronary plaque geometry on plaque vulnerability and its
  association with the risk of future cardiovascular events in patients with
  chest pain undergoing coronary computed tomographic angiography—the
  {GEOMETRY study: Protocol for a prospective clinical trial}.
\newblock Medicine. 2018;97(49).

\bibitem{3067-13}
Morariu M, et~al.
\newblock {Impact of inflammation-mediated response on pan-coronary plaque
  vulnerability, myocardial viability and ventricular remodeling in the
  postinfarction period-the VIABILITY study: Protocol for a non-randomized
  prospective clinical study}.
\newblock Medicine. 2019;98(17).

\end{thebibliography}
	\marginpar{\color{white}E\articlenumber}
\end{document}